\documentclass[lettersize,journal]{IEEEtran}
\usepackage{amsmath,amsfonts}
\usepackage{array}
\usepackage[caption=false,font=normalsize,labelfont=sf,textfont=sf]{subfig}
\usepackage{textcomp}
\usepackage{stfloats}
\usepackage{url}
\usepackage{verbatim}
\usepackage{graphicx}
\usepackage{algorithm}
\usepackage{multirow}
\usepackage{makecell}
\usepackage{changepage}
\usepackage{amsmath}
\usepackage{algpseudocode}
\usepackage{booktabs}
\usepackage{threeparttable}
\usepackage{pifont}
\usepackage{xspace} 
\usepackage{booktabs}
\usepackage{xcolor}
\newcommand{\ie}{\textit{i.e.}\xspace}
\newcommand{\eg}{\textit{e.g.}\xspace}
\newcommand{\etc}{\textit{etc}\xspace}

\def\BibTeX{{\rm B\kern-.05em{\sc i\kern-.025em b}\kern-.08em
    T\kern-.1667em\lower.7ex\hbox{E}\kern-.125emX}}
\usepackage{balance}
\begin{document}
\title{LV-OSD: Language-Vision-Complementary Open-Set Object Detection}

\author{
Yupeng Zhang, 
Ruize Han,
Wei Feng~\IEEEmembership{Member,~IEEE,}
Song Wang~\IEEEmembership{Member,~IEEE,}
Liang~Wan\IEEEauthorrefmark{2}~\IEEEmembership{Member,~IEEE}
\thanks{
% This work was supported by the National Natural Science Foundation of China (Grant No. 62476196), the Guangdong Basic and Applied Basic Research Foundation (Grant No. 2025A1515010101), and the National Natural Science Foundation of China (Grant No. 62402490). 
This work was supported in part by the National Natural Science Foundation of China (Grant Nos. 62476196 and 62402490), and in part by the Guangdong Basic and Applied Basic Research Foundation (Grant No. 2025A1515010101).
(Corresponding authors: Liang Wan.)
}

\thanks{Y. Zhang, W. Feng and L. Wan are with the College of Intelligence and Computing, Tianjin University, Tianjin 300350, China. Email: zhangyupeng@tju.edu.cn, lwan@tju.edu.cn, wfeng@tju.edu.cn.}

\thanks{R. Han, and S. Wang are with the Faculty of Computer Science and Artificial Intelligence, Shenzhen University of Advanced Technology, Shenzhen 518107, Guangdong, China. Email: hanruize@suat-sz.edu.cn, wangsong@suat-sz.edu.cn}
}

% \author{IEEE Publication Technology Department
% \thanks{Manuscript created October, 2020; This work was developed by the IEEE Publication Technology Department. This work is distributed under the \LaTeX \ Project Public License (LPPL) ( http://www.latex-project.org/ ) version 1.3. A copy of the LPPL, version 1.3, is included in the base \LaTeX \ documentation of all distributions of \LaTeX \ released 2003/12/01 or later. The opinions expressed here are entirely that of the author. No warranty is expressed or implied. User assumes all risk.}}

\markboth{IEEE Transactions on Circuits and Systems for Video Technology, IN SUBMISSION}
{How to Use the IEEEtran \LaTeX \ Templates}

\maketitle

\begin{abstract}
Object detection is an important task in computer vision, which aims to detect the objects of interest. through the given category list or query images.
In this work, we propose a new problem of language-visual-complementary open-set object detection (LV-OSD), \textit{i.e.}, using the flexible text-based and/or image-based prompts to specify the desired object categories. This setting is more common and practical in real-world applications.
For this purpose, we design a dual-branch detection framework, \textit{LVDor}, which can simultaneously accept both text and image prompts. Specifically, we first build the Multi-modal Prompts (MPr) containing various text descriptions and image samples for each category. 
Subsequently, to bridge the semantic gap among the input image, text prompts, and image prompts, we design a Target-guided Prompt Dynamic Weighting (TPDW) module. Guided by the prior information of the target image, this module dynamically produces the text and image prompts that best align with the target semantics, achieving precise alignment and effectively reducing the discrepancy between the two modalities, thereby accommodating the LV-OSD setting.
We also propose a simple Prompt Random Masking (PRM) mechanism during training to simulate the arbitrary combination of text and/or image prompts in testing.
Extensive experimental results verify our problem formulation's reasonability and our method's effectiveness.
Prompts and code will be released publicly.
\end{abstract}


\section{Introduction}

\begin{figure}[t!]
	\centering
	\includegraphics[width=0.98\linewidth]{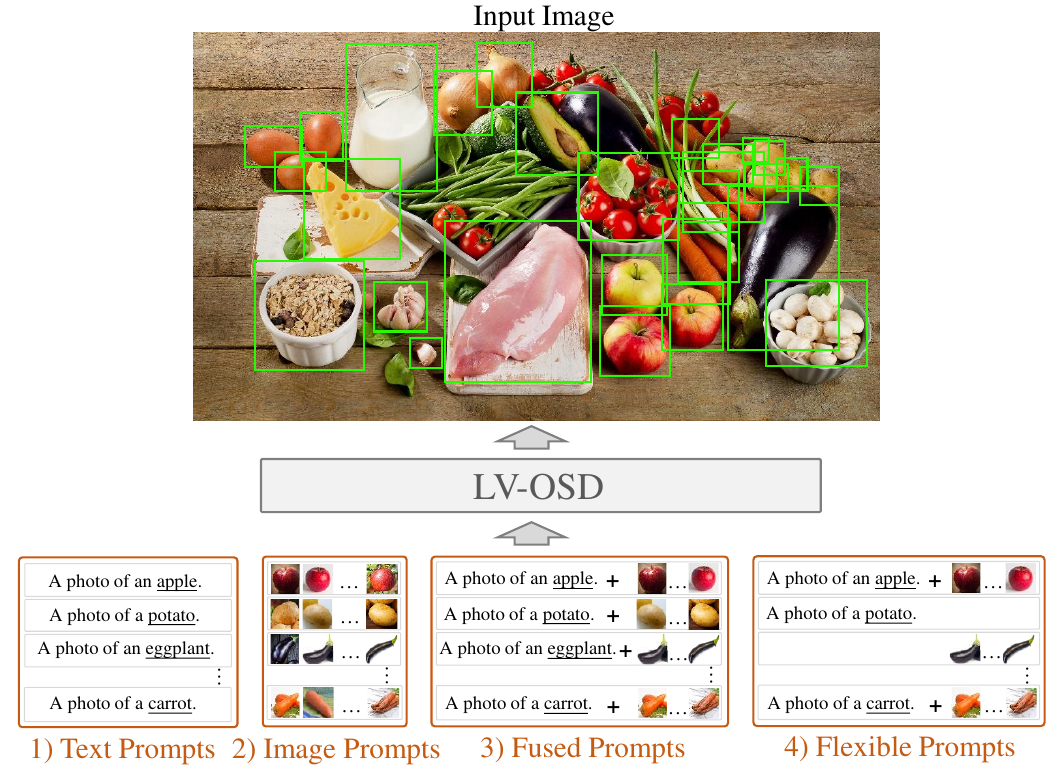}
	\vspace{-10pt}
	\caption{Illustration of the Language-Vision-Complementary Open-Set Object Detection. Our newly proposed task formulation supports four distinct prompt modalities: 1) Text prompts, suitable for scenarios where only category names are available. 2) Image prompts, used when only representative images can be obtained for each category. 3) Fused prompts, applied when both category names and example images are accessible for all categories. 4) Flexible prompts, designed for more realistic cases where some categories provide both names and images, while others offer only one of the two.
    \textbf{Notably, the modality 4) aligns more closely with real-world applications, where the availability of prompt modalities varies across categories, offering greater practicality and flexibility}.} 
	\label{fig:setting}
	\vspace{-15pt}
\end{figure}

Object detection, as a fundamental task in computer vision, has a wide range of practical applications.
The object detection task aims to localize and recognize the objects of interest in a given image.
The desired object categories are commonly specified by a \textbf{text-based query} list.
For example, the classical closed-set object detection, exemplified by the YOLO~\cite{redmon2016you} and Faster RCNN~\cite{ren2016faster} series, is characterized by its ability to recognize only known categories present in the training dataset.
Recently, open-set object detection (OSD)~\cite{zheng2022towards} represented by open-vocabulary object detection (OVD)~\cite{zareian2021open} was proposed to not only identify known categories but also, more importantly, detect entirely new categories unseen during the training phase.
OVD classifies objects by computing the similarity between category text prompts and the features of object regions in the input image.
Current mainstream OVD methods largely rely on manually crafted textual prompts (\eg, `A photo of a dog'). Although these approaches have shown promising results, they still face three key limitations:
First is semantic ambiguities, \eg, confusion from homonyms (`bat' can refer to both the flying mammal and a baseball bat).
Second, the simple text prompts often fail to effectively capture the visual details of objects, leading to the loss of critical visual information. 
Third, in terms of accessibility, the names of certain specific categories are challenging to obtain and accurately convey through text prompts. 
These limitations significantly constrain the adaptability of text-based OVD models in real-world applications.

Besides, \textbf{image-based query} is another means to specify the desired categories; as the saying goes, `a picture is worth a thousand words'.
As representative tasks, few-shot/one-shot object detection (based on metric  learning)~\cite{jiang2023few,wang2024fine,yang2022balanced,zhang2024exploring,su2024toward} aims to identify the open-set categories by measuring the similarity between image examples and the features of regions in the input image. 
The image modality provides richer and more detailed visual cues, effectively compensating for the shortcomings of the text-based category prompt. 
However, image prompts also have limitations. For instance, due to intra-category variation and imaging conditions, a single image cannot fully capture a category’s rich attributes. Furthermore, like text prompts, image prompts face challenges in obtaining representative samples for some categories.

As discussed above, both text and image prompts are useful yet somewhat limited in real applications.
Therefore, in this work, we aim to \textit{explore the complementarity} of such prompts for open-set (real-world) object detection.
Few related works, for example, OV-DETR~\cite{zang2022open} and MM-OVOD~\cite{kaul2023multi} begin to explore incorporating images into text as queries or auxiliary prompts for OVD.
However, \textit{these approaches require uniform prompt formats (whether text, images, or fused representations) for all categories during inference, which significantly increases both operational costs and complexity in practical applications.}
To address this issue, \textbf{we propose a new setting of language-vision-complementary open-set object detection (LV-OSD).
Specifically, LV-OSD enables arbitrary types of prompts combined with language-based text and vision-based images.}
As shown in Fig.~\ref{fig:setting}, in the inference stage, LV-OSD supports the modes of 1) Text prompts, 2) Image prompts, 3) Fused text-image prompts~\cite{zang2022open,kaul2023multi}, and 4) Flexible text-image prompts.
Especially for the new flexible text-image prompts, some categories are guided solely by text-based prompts or by image-based prompts, and the remaining categories utilize both modalities, according to the actual situation.
It can be imagined that this is a more flexible and practical detection method in which, for example, some common categories can use text prompts while some rare categories can use image prompts, and the fused prompts are for the hard categories.

\textbf{We develop a novel object detection framework, LVDor, which is trained jointly with dual-modality prompts.}
This framework aims to equip the detector with the ability to adapt to all types of prompt, as discussed above.
Specifically, on the one hand, for text-modal prompts, we employ a Large Language Model (LLM) to generate multiple web-caption-like descriptions for each category (inspired by CLIP's~\cite{radford2021learning} training data) and utilize a frozen CLIP text encoder to extract embeddings of text prompts.
On the other hand, for image-modal prompts, we curate diverse images from the internet for each category and extract features using a frozen CLIP image encoder to serve as image prompt embedding. 
For each prompt type, the individual prompts are still various, \eg, a category text can generate multiple descriptions using LLM, and the query image of a category also has multiple instances.
Subsequently, to bridge the semantic gap between the input image and multi-modal prompts, we introduce a Target-guided Prompt Dynamic Weighting (TPDW) module. Conditioned on the input image, this module adaptively selects and reweights prompts to achieve precise semantic alignment, thereby exploiting the complementary strengths of different modalities and their flexible combinations. Importantly, TPDW goes beyond simple fusion by enforcing target-aware prompt selection that suppresses semantically inconsistent prompts within the same category, which is crucial under flexible and heterogeneous prompt availability.
Finally, we propose a simple yet effective Prompt Random Masking (PRM) mechanism during training to simulate arbitrary combinations of text and/or image prompts encountered at test time.

%%%%%%%%%%%%%%%%%%%%%%%%%%%%%%%%%%%%%%%%%%%%%%%%%%%%%%%%%%%%%%%%%%%%%%%%%%%%%%%%%%%%%%%%%%%%%%%%%%%%%%%%%%%%%%%%%%%%%%%%

In summary, the main contributions of this work are:
\begin{itemize}
\item We propose a novel and practical language-vision-complementary open-set object detection (LV-OSD) problem, which can be applied across arbitrary prompt scenarios, \ie, text, image, fusion of both, and flexible text-image prompts.

\item We propose a novel object detection framework, termed LVDor, for language–vision complementary open-set object detection (LV-OSD). The framework builds a unified multi-modal prompt set (MPr) by generating diverse textual prompts with large language models and incorporating representative image examples, thereby enhancing category-level semantic coverage. Furthermore, we introduce a Target-guided Prompt Dynamic Weighting (TPDW) mechanism that adaptively selects prompts based on their semantic alignment with the target image to suppress intra-category semantic noise. In addition, an efficient Prompt Random Masking (PRM) strategy is adopted during training to simulate flexible dual-modal prompt combinations, enabling robust adaptation to heterogeneous prompt scenarios.

\item Our method achieves SOTA performance on OV-LVIS and in cross-dataset evaluations on Object365, demonstrating the proposed framework's strong robustness and generalization ability for open-set object detection under multi-modal prompts.
\end{itemize}

\section{Related Work}
\label{sec:Related work}
\textbf{Text-prompt-based detection} includes traditional closed-set detection, such as the Faster RCNN~\cite{ren2016faster}, where detectors recognize only the categories they were trained on. A more researched area is OVD, which detects categories not seen during training without requiring model retraining, falling under open-set detection~\cite{zheng2022towards}. The emergence of VLMs, such as CLIP, have advanced OVD. These models, by jointly learning image and text representations, support OVD. Some studies~\cite{kim2023contrastive,kim2023detection,kim2023region,wu2023cora,song2023prompt} use region-aware training methods, integrating image-text pairs into detection training to improve classification for novel categories. Other approaches~\cite{zhou2022detecting,zhong2022regionclip,jeong2024proxydet,ma2024codet,kaul2023multi} leverage pre-trained VLMs or generate pseudo-labels to train detectors. Additionally, methods like~\cite{gu2021open,pham2024lp,bangalath2022bridging,li2023distilling,du2022learning} use knowledge distillation to provide detectors with open-vocabulary capabilities. Furthermore, some research~\cite{kuo2022f,minderer2022simple,wu2023clipself,xu2024dst,wang2025declip} builds open-vocabulary detectors on frozen pre-trained models to exploit the generalization power of large-scale models.

\textbf{Image-prompt-based detection} primarily focuses on Few-Shot (FS) and One-Shot (OS) object detection. FSOD predominantly follows two technical approaches. The first is the fine-tuning method, as demonstrated by~\cite{liu2023integrally,wang2024snida,10007034}, which adjusts network parameters using limited data from novel categories to detect novel objects. While effective, these methods are unsuitable for OVD, as they necessitate retraining or fine-tuning. The second approach employs Siamese networks, as seen in~\cite{jiang2023few,wang2024fine,su2024toward,10298598,9600874}, which process target and query images in parallel, measuring the similarity between image regions (typically bounding boxes) and few-shot queries (prompts) to detect both base and novel categories.
OSOD~\cite{yang2022balanced,zhang2024exploring}, a special case of FSOD, utilizes only a single image example and adopts the same methodology. However, the dual-branch architecture in FSOD lacks text prompts, thereby limiting its practicality and flexibility. 

\textbf{Image and Text Prompts Fusion for Object Detection.} OV-DETR~\cite{zang2022open} and MM-OVOD~\cite{kaul2023multi} highlight the limitations of relying solely on text prompts in OVD by incorporating image modalities as complementary cues. OV-DETR adapts the DETR framework using a frozen CLIP to extract image features for inference, though it only offers qualitative results. MM-OVOD also employs a frozen CLIP to extract multi-modal features and trains an offline image aggregation network to fuse multiple image examples. However, its rigid alignment between image and text prompts leads to the loss of fine-grained visual information. Both methods overlook the modality gap and the complementary strengths of image and text prompts~\cite{liang2022mind}.
To address this, we propose a novel LV-OSD framework. Unlike prior work, our method uses a frozen CLIP image encoder as the backbone, preserving pretrained knowledge without costly retraining. By jointly training with image and text prompts, our approach enables flexible adaptation to both modalities.

\begin{figure*}[t!]\vspace{-10pt}
	\centering
	\includegraphics[width=0.95\linewidth]{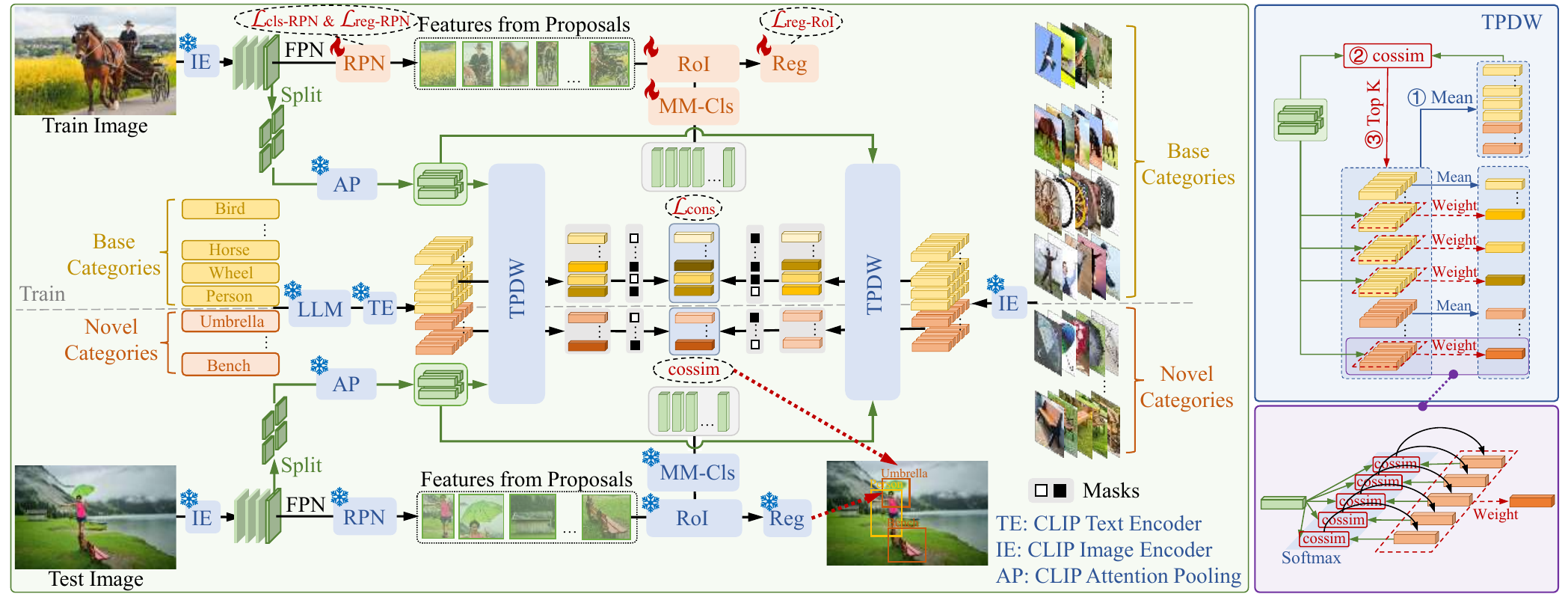}
	\vspace{-10pt}
	\caption{Overview of our method. We employ a LLM to generate category descriptions and collect representative images to construct a comprehensive set of text and image prompts. Frozen CLIP text and image encoders serve as the backbone of our framework. During training, we leverage a Target-guided Prompt Dynamic Weighting (TPDW) method to select and fuse the most semantically relevant prompts based on the input image. A Prompt Random Masking (PRM) mechanism is introduced to simulate flexible prompt combinations, enhancing the model's adaptability to various input forms. At inference, the TPDW similarly selects and fuses the most relevant prompts from the available multi-modal prompts to guide open-set object detection.} 
    \label{fig:overall}
	\vspace{-10pt}
\end{figure*}

\begin{figure}[t!]\vspace{-7pt}
	\centering 
	\includegraphics[width=1.0\linewidth]{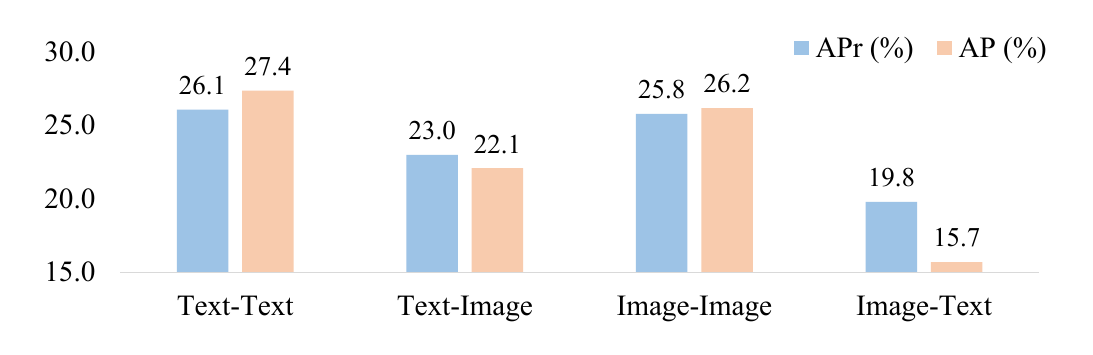}
	\vspace{-28pt}
	\caption{Illustration of the modality gap in OVD with different modality prompts. `A-B' denotes the modality of `A' for training and the modality of `B' for testing.} 
	\label{fig:modility gap}
	\vspace{-18pt}
\end{figure}

\section{Proposed Method}
\label{sec:Method}
\subsection{Preliminaries and Framework}
\textbf{Problem formulation.}
Given a target image \( \mathbf{I} \in \mathbb{R}^{3 \times H \times W} \) as input to the detector, two types of outputs are typically desired: 
\textbf{(1) Location}, where the bounding box coordinates \( \mathbf{b}_i \in \mathbb{R}^4 \) represent the location of the \( i \)-th predicted object. 
\textbf{(2) Category}, where \( P_i^j \in \mathcal{C}_{\text{test}} \) representing the \( j \)-th category is assigned to $\mathbf{b}_i$, with \( \mathcal{C}_{\text{test}} \)  as the set of categories during testing.
Following the open-set setting, in the training phase, the categories consist solely of the set of base category \( \mathcal{C}_{\text{base}} \), while during testing, the vocabulary is extended to include the set of novel category names \( \mathcal{C}_{\text{novel}} \), \ie, $\mathcal{C}_{\text{test}} = \mathcal{C}_{\text{base}} \cup \mathcal{C}_{\text{novel}}$ with \( C_{\text{base}} \cap \mathcal{C}_{\text{novel}} = \emptyset \).

\textbf{The framework.} We explore the use of a frozen VLM as the backbone for LVDor, training only the detection head to significantly reduce training complexity while avoiding the need for knowledge distillation or task-specific pretraining. 
We begin by designing four training--testing prompt combinations and conducting preliminary experiments:
\begin{itemize}
    \item training the detection head with text-modal prompts and testing with text (T-T) and image prompts (T-I);
    \item training the detection head with image-modal prompts and testing with text (I-T) and image prompts (I-I).
\end{itemize}
As shown in Fig.~\ref{fig:modility gap}, the results show that mismatched prompt modalities between training and testing lead to a substantial performance drop. This phenomenon arises from the modality-gap issue discussed in~\cite{liang2022mind} and can also be attributed to the asymmetric representations induced by the inherent diversity of the two prompt modalities; under the flexible-prompt setting, this asymmetry is further amplified and severely impairs model stability.
To mitigate this issue, we retain both the text and image branches in the framework and introduce a Target-guided Prompt Dynamic Weighting (TPDW) module, which dynamically fuses the embeddings of the two modal prompts. Compared with the direct fusion strategy in~\cite{kaul2023multi}, TPDW more effectively bridges the semantic gap between modalities, producing fused representations better aligned with the target object.
The fused multi-modal prompt embeddings are then used to train a multi-modal classification (MM-Cls) head and aligned with RoI features. Although MM-Cls is a standard MLP, within our framework it can simultaneously align with both text- and image-modal prompts while maintaining consistent discriminative capability across modalities.

\textbf{Overview.} In general, as shown in the left of Fig.~\ref{fig:overall}, we employ frozen CLIP image and text encoders as the foundational network. 
First, to enrich the arbitrary multi-modal prompts for the open-set object detection, we first utilize a DeepSeek~\cite{guo2025deepseek} to generate multiple category descriptions (text prompt) and crawl a certain number of image examples (image prompt) from the internet for each category, forming the diverse Multi-modal Prompts (MPr) for training.
Subsequently, to narrow the semantic gap among the input image, text prompts, and image prompts and to effectively train the detection head, we propose the Target-guided Prompt Dynamic Weighting (TPDW) module. This module selects the top $K$ most semantically relevant categories based on the similarity between the input image and all prompts, and then adaptively fuses the multiple prompts within each selected category according to their alignment with the image, thereby producing refined prompt representations.
Finally, by introducing a Prompt Random Masking (PRM) mechanism, we generate arbitrary multi-modal prompts to train the classification head to handle the uncertainty of given image and/or text prompts during testing. 

\subsection{Construction of Multi-modal Prompts (MPr)}
\label{sec:prompts}
\textbf{Text-based category description generation.}
Recent studies~\cite{wu2023clipself,bangalath2022bridging} on CLIP-based OVD manually design prompt templates based on dataset characteristics, such as `a black-and-white photo of \{\}' or `a photo of \{\} in the rain'. However, these prompts are easy to obtain but fail to leverage the potential of VLMs, limiting generalization across datasets. Moreover, due to intra-category variability, objects in the same category may exhibit diverse visual attributes. Thus, \textit{a fixed prompt for each category inevitably leads to insufficient generalization}.

To address these challenges, we propose a diversified text-prompt generation strategy based on LLMs. Specifically, we employ LLMs to generate $N$ ($N=5$) category-specific descriptions for each category in the dataset, written in the style of internet image captions and enriched with fine-grained attributes that distinguish visually similar categories, thereby producing more detailed and discriminative semantic representations and enhancing the model's generalization ability.
To this end, we use DeepSeek to generate category descriptions and provide the following instruction template for each category: \textit{`Please generate feature descriptions for the following categories in the style of internet image captions, emphasizing the diversity of their visual attributes.'} During generation, we emphasize both the salient semantic characteristics of each category and the distinctiveness among the descriptions, ensuring that the five descriptions cover a wide range of representative scenarios. For example: `An airplane has a long, cylindrical fuselage with wings on either side and a tail at the rear', `An airplane typically features a pointed nose, a cockpit at the front, and windows along the fuselage', \etc.
In this way, each description not only effectively distinguishes the category from others but also captures intra-category diversity by highlighting different visual aspects such as shape and texture. Based on this strategy, we construct rich and varied category-specific text prompts that guide the model toward a more comprehensive and robust semantic understanding of each category, enabling it to better handle substantial intra-category variation in real-world detection scenarios.

\textbf{Image-based category example collection.}
In addition to textual descriptions for text prompt, leveraging image examples serves as a natural and effective complementary strategy, particularly when category descriptions are imprecise or category names are occasionally unavailable. 
This way, we collect a diverse set of example images from the Internet for each category. 
During collection, we place particular emphasis on diversity, encompassing various factors such as lighting conditions, shooting angles, background environments, object poses, scene complexity, \etc.
For instance, for the `airplane' category, we gather images from diverse scenarios, including daytime and nighttime, takeoff and landing, as well as close-up and distant views. 
Furthermore, we ensure the example images cover a range of resolutions, clarity levels, and styles (such as real photographs and artistic renderings), thereby enhancing the model's adaptability to produce diverse visual features. 
Through this diversified image collection strategy, we aim to improve the model's robustness and generalization capabilities when confronted with complex real-world scenarios.

\subsection{Target-guided Prompt Dynamic Weighting (TPDW)}
\label{sec:TPDW}
Intra-category diversity causes the collected textual descriptions and example images of the same category to exhibit substantial differences: concise textual prompts often convey diverse high-level semantics, whereas example images—containing richer visual information—vary widely in viewpoint, shape, color, illumination, and other appearance factors. This modality-specific diversity creates an inherent asymmetry in their representation spaces, which in turn leads to two critical issues. First, during training, semantic inconsistencies between prompts and target objects in the input image may induce mismatches and hinder effective learning. Second, under the Flexible Prompts evaluation setting, this asymmetry becomes further amplified, significantly degrading the accuracy of category matching.

To this end, we propose the \textbf{Target-guided Prompts Dynamic Weighting (TPDW)} module. TPDW dynamically assigns weights to multiple text and image prompts for each category based on the semantic information of the input image, enabling their adaptive fusion. 
As shown on the right side of Fig.~\ref{fig:overall}, TPDW in the training phase adaptively aligns prompts with the semantic characteristics of the objects in the input train image, avoiding forced alignment with prompts of the same category that are semantically irrelevant. 
During testing, TPDW similarly adjusts prompt weights according to the content of the test image, further improving the accuracy and robustness of feature matching.

Specifically, we extract prompt embeddings for the $N$ textual descriptions and the example images collected for each category using the frozen CLIP text and image encoders, respectively.
We then extract the last-layer features from the input image feature map as the `reference' for prompt adjustment.
\textit{To reduce computational overhead}, we divide the input image into four patches in the feature space, with each patch independently guiding the dynamic weighting process.
Taking text prompts as an example, given the feature of an image patch $p$, \ie, $\mathbf{f}_p$, we first compute the cosine similarity between it and the average prompt embeddings of each category as
\begin{equation}
\footnotesize
\label{eq:1}
s^p_c = \frac{\mathbf{f}_p \cdot \bar{\mathbf{e}}_c}{\|\mathbf{f}_p\| \cdot \|\bar{\mathbf{e}}_c\|},
\end{equation}
where $\bar{\mathbf{e}}_c$ is the average prompt (of $N$ prompts) embedding vector for category $c$, ${s^p_c}$ is the cosine similarity between the patch and category $c$.
With $s^p_c$ for all $c \in \mathcal{C}$, $\mathcal{C}$ is the set of all categories, and we roughly estimate the potential target categories within this patch.
Next, we select the top $K$ categories with the highest similarity as candidate target categories as
\begin{equation}
\footnotesize
\label{eq:2}
\mathcal{C}_{\text{top $K$}} = {\arg\max}_K s^p_c, c \in \mathcal{C},
\end{equation}
where $\mathcal{C}_{\text{top $K$}}$ is the set of top $K$ candidate categories.
Since the CLIP image encoder is frozen, this rough estimation is highly reliable.

Then, we utilize the features of the patch $\mathbf{f}_p$ to compute the cosine similarity with the $N$ text prompts for each of the $K$ categories as
\begin{equation}
\footnotesize
\label{eq:3}
w_c^i = \frac{\exp(\text{cossim}({\mathbf{f}_p, \mathbf{e}_c^i}))}{\sum_{n=1}^{N}\exp(\text{cossim}({\mathbf{f}_p, \mathbf{e}_c^i}))},
\end{equation}
where \text{cossim} means the cosine similarity, $p_c^i$ is the $i$-th prompt vector for category $c$, $w_c^i$ is the weight for the $i$-th prompt of category $c$.
These similarity values are then transformed into weights via the softmax function, and the corresponding prompts are weighted accordingly. 
The prompt embeddings are re-assigned with these weights as
\begin{equation}
\footnotesize
\label{eq:4}
\mathbf{e}^{\text{weight}}_c = \sum_{i=1}^{N} w_c^i \cdot \mathbf{e}_c^i.
\end{equation}

The same procedure is applied to the remaining patches.
Finally, the weighted prompt results from the all patches are integrated. 
If a category $c$ is weighted, \ie, $|\mathcal{P}_c| > 0$, the average of the weighted prompts is taken as the final prompt for that category. 
For categories not re-weighted, the average of their $N$ prompts is used as the final prompt as
\begin{equation}
\footnotesize
\label{eq:5}
\mathbf{e}^{\text{final}}_c = 
\begin{cases} 
\frac{1}{|\mathcal{P}_c|} \sum_{p \in \mathcal{P}_c} \mathbf{e}^{\text{weight}}_c (p) & \text{if } |\mathcal{P}_c| > 0 \\ 
\frac{1}{N} \sum_{i=1}^{N} \mathbf{e}_c^i & \text{otherwise}
\end{cases},
\end{equation}
where $\mathcal{P}_c$ is the set of patches that re-weighted category $c$, and $\mathbf{e}^{\text{final}}_c$ is the final prompt for category $c$.

The same procedure is applied to the image prompts as to the text prompts.
Through this approach, we dynamically adjust prompts more precisely based on the content of the input image, thereby enhancing the model's performance in LV-OSD tasks.

\subsection{Prompt Random Masking (PRM) Mechanism} 
\label{sec:mask}
The final step is to combine the text-based and image-based prompt branch, 
As discussed before, the proposed framework is required to receive the arbitrary-combined multi-modal prompts at inference, \ie, only text or image prompts or the flexible text-image prompts.
This way, during training, to simulate the arbitrary multi-modal prompts, we introduce a Prompt Random Masking mechanism. Specifically, as shown in the left of Fig.~\ref{fig:overall}, we added the mask into the text prompts $\mathbf{P}^{\text{Text}}$ and image prompts $\mathbf{P}^{\text{Img}}$ as 
\begin{equation}
\label{eq:mask-text}
\footnotesize 
\hat{\mathbf{P}}^{\text{Text}} = \mathbf{P}^{\text{Text}} \odot \mathbf{M}^{\text{Text}},
\hat{\mathbf{P}}^{\text{Img}} = \mathbf{P}^{\text{Img}} \odot \mathbf{M}^{\text{Img}},
\end{equation}
where $\hat{\mathbf{P}}^{\text{text}} \in \mathbb{R}^{d\times C}$ represents the masked text prompt embeddings, $\hat{\mathbf{P}}^{\text{img}} \in \mathbb{R}^{d\times C}$ denotes the masked image prompt embeddings, and $\mathbf{M}^{\text{text}}, \mathbf{M}^{\text{text}} \in \{0,1\}^C$ denote the mask for text and image prompts, respectively, $C$ represents the number of categories.

This mechanism randomly masks one of the two modalities for post-fusion.
The two concurrent prompts from both modalities are added. After the normalization, the multi-modal aggregated prompt for each category is generated. 
This process enables the final prompt list to simulate the aforementioned scenarios under complex application environments involving uncertain multi-modal prompts. 
This way, we generate various combinations of flexible multi-modal prompts.

\subsection{Framework and Details}

\textbf{Training stage.} 
The total loss function includes 
\begin{equation}
\label{eq:total_loss}
\footnotesize 
\mathcal{L}_{\text{total}} = \mathcal{L}_{\text{reg-RPN}} + \mathcal{L}_{\text{cls-RPN}} + \mathcal{L}_{\text{reg-RoI}} + \mathcal{L}_{\text{cons}},
\end{equation} 
where $\mathcal{L}_{\text{cls-RPN}}$, $\mathcal{L}_{\text{reg-RPN}}$ and $\mathcal{L}_{\text{reg-RoI}}$ are the classical classification and regression losses for detection, and $\mathcal{L}_{\text{cons}}$ is the contrastive loss for the MM-Cls head. 
We apply the optimized CLIP ViT-B/16 and ViT-L/14 from CLIPSelf~\cite{wu2023clipself} and its improved version DeCLIP~\cite{wang2025declip} as our backbones, During training, we keep the backbone (image and text encoders) frozen, and only train the detection heads. Simultaneously, we interpolate the feature maps from layers $[3, 5, 7, 11]$ of ViT-B/16 with relative scales $\left[\frac{1}{4}, \frac{1}{8}, \frac{1}{16}, \frac{1}{32}\right]$ to the input image size. For ViT-L/14, we interpolate the feature maps from layers $[6, 10, 14, 23]$ with relative scales $\left[\frac{1}{3.5}, \frac{1}{7}, \frac{1}{14}, \frac{1}{28}\right]$ to the input image size. 
We train the model for 50 epochs on the OV-LVIS. We use 16 NVIDIA 3090 GPUs, with a batch size of 10 per GPU, and we use the AdamW optimizer with a learning rate of \(10^{-4}\) and a weight decay of 0.1.
During training, we utilize prompts from both modalities and the proposed random masking mechanism to train the MM-Cls head.

\textbf{Testing stage.} As shown in the left of Fig.~\ref{fig:overall}, we collect five text prompts and five image prompts for each \textit{base} and \textit{novel} category, and employ TPDW to obtain image-adaptive prompt representations for the test image. Finally, we localize objects using proposals generated by the RPN, and classify both base and novel objects by computing the cosine similarity between proposal features and the adjusted prompts from different modalities and their combinations.

\section{Experiments}
\subsection{Setup}
\textbf{Datasets and evaluation metrics.}
Our method is evaluated on the standard OVD benchmark, LVIS. LVIS comprises 100K images and 1,203 categories. The categories are divided into three groups based on the number of training images: `frequent', `common', and `rare'. Following ViLD~\cite{gu2021open}, we treat 337 `rare' categories as \textit{novel categories} and train exclusively on the \textit{base categories} (405 `frequent' and 461 `common' categories). This benchmark is referred to as OV-LVIS. We follow previous work in reporting the average precision for OV-LVIS. we report the average precision for `frequent', `common', and `rare' categories, denoted as $AP_f$, $AP_c$, and $AP_r$ respectively. The symbol $AP$ represents the average precision across all categories. 

\begin{table}[t!]
\setlength{\tabcolsep}{3.2pt}
	\centering
	% \scriptsize
        \tiny
        \caption{Comparison with SOTA methods on OV-LVIS.} \vspace{-5pt}
	    \label{tab:LVIS}
        \begin{threeparttable}
	    \begin{tabular}{l|c|c|c|c|c|c|c}
        \toprule
        Method    & Backbone & Modality & Train Data  &$AP_f$     &$AP_c$    &$AP_r$  &$AP$  \\\hline
        \multirow{2}*{RegionCLIP} & {RN50}$^*$ & \multirow{2}*{T}   & \multirow{2}*{CC3M}  & 34.0 & 27.4 & 17.1 &  28.2 \\
         & {RN50x4}$^*$ &  &   & 36.9 & 32.1 & 22.0 & 32.3 \\
        \hline
        Detic  & {RN50}$^*$ & T & LVIS-base + IN-L & -  & -  &24.9  &32.4\\
        \hline
        \multirow{2}*{OWL-ViT} & ViT-B/16 & \multirow{2}*{T} & \multirow{2}*{O365 + VG}  & - & - & 20.6 &  27.2 \\
         & ViT-L/14 &  &  & - & - & 31.2 & 34.6 \\
        \hline
        \multirow{4}*{RKDWTF} & {RN50}$^*$ Base & \multirow{4}*{T}  & \multirow{4}*{LVIS-base + IN-L}  & 26.4 & 19.4 & 12.2 & 20.9 \\
         & {RN50}$^*$ RKDPIS &  &  & 25.5 & 20.9 & 17.3 & 22.1 \\
         & {RN50}$^*$ WTF &  &  & 26.7 & 21.4 & 17.1 & 22.8 \\
         & {RN50}$^*$ WTF8x &   &  & 29.1 & 25.0 & 21.1 & 25.9 \\
        \hline
        CORA & \multirow{2}*{RN50x4}  &\multirow{2}*{T}  &LVIS-base &-  &-  &22.2 &-  \\
        \cline{1-1} \cline{4-4}
        CORA+ &  & &LVIS-base + IN-21K   &-  &-  &28.1 &-  \\
        \hline
        OADP & ViT-B/32 & T & LVIS-all  & 32.0 & 28.4 & 21.9 & 28.7 \\ 
        \hline
        DK-DETR & RN50 & T  & LVIS-all  & 40.2 & 32.0 & 22.2 & 33.5 \\
        \hline
         YOLO-World & {YOLOv8-L}$^*$ & T  & \multirow{2}*{O365 + GoldG}  &35.4 &24.9 &22.9 &28.7 \\
         \cline{1-3}\cline{5-8}
        YOLOE & {YOLOv11-L}$^*$  & T &  &36.5 &35.0 &29.1 &35.2 \\
        \hline
        \multirow{6}*{MM-OVOD} & \multirow{6}*{{RN50}$^*$} & T  & \multirow{3}*{\makecell{LVIS-base}} & - & - &19.3  &30.3  \\
         &  & I  &  & - & - &18.3  &29.2  \\
         &  & F  &  & - & - & 19.3 & 30.6 \\
        \cline{3-8}
         &  & T  &  \multirow{3}*{\makecell{LVIS-base + IN-L}}  & - & - &25.8  &32.7  \\
         &  & I  &    & - & - &23.8  &31.3  \\
         &  & F  &   & - & - & 27.3 & 33.1 \\
        \hline
        DST-Det & ViT-B/16 &\multirow{6}*{T} & \multirow{32}*{LVIS-base} & -  & -  & 26.2  & -  \\
        \cline{1-2}\cline{5-8}
        \multirow{2}*{SAS-Det} & RN50-C4 & &  &31.6  & 26.1 & 20.9  &27.4 \\
	       & RN50x4-C4 &  &  &36.8  &32.4  &29.1  &33.5 \\
        \cline{1-2}\cline{5-8}
        LBP & - & &  &32.4  &28.8   &22.2 &29.1  \\
        \cline{1-2}\cline{5-8}
        \multirow{2}*{OV-DQUO} & ViT-B/16 & & &23.8  &27.7   &29.4  &26.5  \\
        & ViT-L/14 & & &28.5  &36.0  &39.5  &33.7  \\ 
        \cline{1-3}\cline{5-8}
        \multirow{4}*{F-VLM (CLIP)} & RN50 & \multirow{4}*{T}  &  & - & - & 18.6 & 24.2 \\
         & RN50x4 &  &  & - & - & 26.3  & 28.5 \\
         & RN50x16 &  &  & - & - & 30.4 & 32.1 \\
         & RN50x64 &  &  & - & - & 32.8 & 34.9 \\
         \cline{1-3}\cline{5-8}
         \multirow{6}*{\textbf{LVDor} (CLIP)}  & \multirow{3}*{RN50} & T  &  &27.1  &26.2  &22.3  &25.9  \\
         &  & I &  & 26.8  & 26.3  & 21.4  & 25.7 \\
         &  & F &  & 27.3  & 26.6 &22.8  & 26.2 \\
         \cline{2-3}\cline{5-8}
         & \multirow{3}*{RN50x4} & T &  &30.4  &29.7  &28.8  &29.8  \\
         &  & I &  &29.9  &29.5  &27.9  &29.4  \\
         &  & F &  &30.6  &30.1  &28.9  &30.1  \\
        \cline{1-3}\cline{5-8}
    	\multirow{2}*{F-ViT (CLIPSelf)} & ViT-B/16 & \multirow{2}*{T}  &   &29.1  &21.8 &25.3 &25.2  \\
    	  & ViT-L/14 &  &  &35.6  &34.6  &34.9  &35.1  \\
        \cline{1-3}\cline{5-8}
        \multirow{6}*{{\textbf{LVDor} (CLIPSelf)}} & \multirow{3}*{ViT-B/16} & T & &31.4  &24.6  &29.6  &28.1   \\
         &  & I &  &30.7   & 27.2 & 26.4 & 28.4  \\
         &  & F &  &31.5   &26.8  &29.8  &29.2   \\
         \cline{2-3}\cline{5-8}
         & \multirow{3}*{ViT-L/14} & T  &  &38.0  & 36.2 & 39.8 &37.5   \\
         &  & I  &  & 37.5  & 36.3 & 37.2 & 36.9  \\
         &  & F &  &37.8  &36.7  & 39.8  &37.7  \\
         \cline{1-3}\cline{5-8}
    	\multirow{2}*{F-ViT (DeCLIP)} & ViT-B/16 & \multirow{2}*{T}  &   &29.8  &22.4  &26.8 &26.0  \\
    	  & ViT-L/14 &  &  &36.5  &35.2  &37.2  &36.0  \\
        \cline{1-3}\cline{5-8}

        \multirow{6}*{{\textbf{LVDor} (DeCLIP)}} & \multirow{3}*{ViT-B/16} & T  &  &31.9  &25.3  &30.4  & 28.8  \\
         &  & I &  &31.4  &28.1  &26.9  &29.2  \\
         &  & F &  &32.3  &28.2  &30.7  &30.2    \\
         \cline{2-3}\cline{5-8}
         & \multirow{3}*{ViT-L/14} & T  &  &38.2  &37.1  &40.3  &38.1   \\
         &  & I  &  &37.9  &36.9  &38.4  &37.5  \\
         &  & F &  &38.6  &37.3  &40.9  &38.4   \\
        \bottomrule    
	    \end{tabular}
        \begin{tablenotes} 
        % \footnotesize
        \scriptsize
        \item Notes: IN-L denotes the inclusion of images corresponding to the 997 categories shared between ImageNet-21k-P (IN-21K)~\cite{ridnik2021imagenet} and LVIS, `$^*$' indicates that the backbone is not initialized with CLIP. CC3M~\cite{sharma2018conceptual}, GoldG~\cite{kamath2021mdetr}, VG~\cite{krishna2017visual}, and ALIGN~\cite{jia2021scaling} are all publicly datasets.
        \end{tablenotes} \vspace{-15pt}
\end{threeparttable}
\end{table}

\textbf{Comparison methods.} 
For a fair comparison, we select several mainstream VLM-based OVD methods for testing, evaluation, and comparison on the OVD benchmarks. 
Specifically, we include the transfer learning approaches, \ie, F-VLM~\cite{kuo2022f}, CLIPSelf~\cite{wu2023clipself}, DeCLIP~\cite{wang2025declip}, DST-Det~\cite{xu2024dst}, LBP~\cite{li2024learning}, OV-DQUO~\cite{wang2025ov} and OWL-ViT~\cite{minderer2022simple}, and several knowledge distillation methods, \ie, OADP~\cite{wang2023object}, RKDWTF~\cite{bangalath2022bridging}, DK-DETR~\cite{li2023distilling}, RegionCLIP~\cite{zhong2022regionclip}, pseudo-labeling method MM-OVOD~\cite{kaul2023multi}, Detic~\cite{zhou2022detecting}, SAS-Det~\cite{zhao2024taming}, and region-aware training method RO-ViT~\cite{kim2023region}, CORA~\cite{wu2023cora}, YOLO-World~\cite{cheng2024yolo} and YOLOE~\cite{wang2025yoloe}, which retrain a network from scratch using large-scale datasets.
We adopt the optimized CLIP ViT-B/16 and ViT-L/14 from CLIPSelf and its improved version DeCLIP as our backbones and apply our method on top of them, using the F-ViT proposed in CLIPSelf as the baseline detection framework.
In addition, we integrate our method into F-VLM as the baseline framework and further validate its effectiveness using CLIP's RN50 and RN50×4 as backbones.

\subsection{Main Results on OV-LVIS}
Table~\ref{tab:LVIS} presents the experimental results of various methods on OV-LVIS. 
Since most methods rely solely on the text modality for detection, we first compare the text-based results (`T'). 
As shown in the table, regardless of whether CLIP, CLIPSelf, or DeCLIP is used as the backbone, our LVDor consistently achieves the best performance. 
We also observe that using the image modality (`I') as a prompt yields strong results, further demonstrating its effectiveness in detection. 
Moreover, when fusing both text and image modalities (`F'), the model attains substantially higher performance across all metrics than using either modality alone, highlighting the importance of multimodal fusion for open-vocabulary detection. 
It is worth noting that our method is trained solely on LVIS-base. Despite other methods training the detection framework on much larger datasets (\eg, OWL-ViT) or further pretraining the visual backbone on even broader data (\eg, YOLO-World, YOLOE, MM-OVOD), our model still delivers competitive performance. 
These results collectively demonstrate the effectiveness of LVDor in enhancing generalization for open-set and open-vocabulary detection.

\subsection{Results using Flexible Multi-Modal Prompts} 
We validate the proposed novel task setting—flexible multi-modal prompts detection, on the OV-LVIS dataset. 
We provide three settings to simulate flexible multi-modal scenarios: 
(1) we randomly select half of the categories from both base and novel categories and provide only text-based prompts, while the other half receive image-based prompts (denoted as T/2-I/2), 
(2) only image for one half of the categories and `text + image' for the other half (denoted as T-I/2); 
(3) only text for one half of the categories and the other half for both modalities (denoted as T/2-I). 
In this way, we construct three mixed-modal prompt scenarios. We then compare and evaluate MM-OVOD, F-VLM (CLIP), F-ViT (CLIPSelf and DeCLIP), and our proposed method LVDor (CLIP, CLIPSelf, and DeCLIP).

\setlength{\tabcolsep}{3.2pt}
\begin{table}[t!] 
	\centering
	\tiny
    % \scriptsize
    % \footnotesize
    \caption{Comparison of results using \textbf{flexible multi-modal prompts} on the OV-LVIS.} \vspace{-5pt}
	\label{tab:mix-LVIS}
	\begin{tabular}{l|c|c|c|c|c|c|c}
        \toprule
        Method    & Backbone &Modality & Train Data  &$AP_f$     &$AP_c$    &$AP_r$  &$AP$  \\\hline
        \multirow{6}*{MM-OVOD} &\multirow{6}*{{RN50}$^*$} &T/2-I/2  &\multirow{3}*{LVIS-base}  &28.7  &21.5 &11.7 &22.7 \\
         & &T-I/2 & & 30.4  & 22.6  & 12.7 & 24.0 \\
         & & T/2-I & & 28.9 & 23.2 & 12.2 & 23.5 \\
        \cline{3-8}
         &  &T/2-I/2 &\multirow{3}*{LVIS-base + IN-L}   & 29.8  & 24.2   & 15.2  & 24.8  \\
         &  &T-I/2 & & 31.5  & 25.6  & 16.7  &26.4  \\
         &  & T/2-I &  & 30.6  & 27.5  & 19.1  & 27.3  \\
        \hline 
	   \multirow{6}*{F-VLM (CLIP)} & \multirow{3}*{RN50} &T/2-I/2 &\multirow{37}*{LVIS-base}  &23.2  & 19.9 &17.8  &20.8  \\
         &  &T-I/2 &  &22.6  &23.0  &18.8  &22.1  \\
         &  & T/2-I &  &23.0  &22.7  &17.9  &22.0 \\
        \cline{2-3}\cline{5-8}
	   & \multirow{3}*{RN50×4} &T/2-I/2 &  &26.7  &22.0  &23.8  &24.1 \\
         &  &T-I/2 &  &25.9  &22.3  &25.6  & 24.3 \\
         &  & T/2-I &  &25.2  &22.4  &24.4  & 23.8 \\
        \cline{1-3}\cline{5-8}
        \multirow{6}*{\textbf{LVDor} (CLIP)} & \multirow{3}*{RN50} &T/2-I/2 &  &26.1  &23.8  &19.6  &24.0  \\
         &  &T-I/2  &  &25.8  &25.1  &19.9  & 24.5  \\
         &  & T/2-I &  &25.6  &25.2  &20.3  &24.5 \\
        \cline{2-3}\cline{5-8}
         & \multirow{3}*{RN50×4} &T/2-I/2 &  &29.2  &25.9  &26.7  & 27.3  \\
         &  &T-I/2   &  &29.8  &25.2  &28.8  & 27.6 \\
         &  & T/2-I  &  &29.5  &25.7  &27.5  & 27.5 \\ \cline{1-3}\cline{5-8}
	   \multirow{6}*{F-ViT (CLIPSelf)} & \multirow{3}*{ViT-B/16} &T/2-I/2 &   &25.6  &20.7 &13.6 &21.4  \\
         &  &T-I/2 &  &28.9  &21.6  &15.6  &23.4  \\
         &  & T/2-I &  & 25.5 &22.9  &22.7  &23.9  \\
        \cline{2-3}\cline{5-8}
	   & \multirow{3}*{ViT-L/14} &T/2-I/2 &  &31.5 &33.0  &17.3  &29.7  \\
         &  &T-I/2 &  &33.1   &34.7  &21.4  &31.8  \\
         &  & T/2-I &  & 31.8  & 32.8 &33.6  &32.6 \\
        \cline{1-3}\cline{5-8}
        \multirow{6}*{\textbf{LVDor} (CLIPSelf)} & \multirow{3}*{ViT-B/16} &T/2-I/2 &  &29.9  &26.2  &27.1  &27.8  \\
         &  &T-I/2  &  & 30.8  &25.8  & 29.0 & 28.3   \\
         &  & T/2-I &  &30.2   & 26.2 & 28.4 & 28.1 \\
        \cline{2-3}\cline{5-8}
         & \multirow{3}*{ViT-L/14} &T/2-I/2 &  &37.2  &36.4  &37.5  &36.9   \\
         &  &T-I/2   &  & 37.6 &36.5  & 38.8 & 37.3  \\
         &  & T/2-I  &  & 37.4 & 36.1 & 38.4 & 37.0   \\ \cline{1-3}\cline{5-8}
        {\textbf{LVDor} (CLIPSelf) \textbf{w/o PRM}}& {ViT-B/16} &T/2-I/2  &  & 28.9  &25.8  & 26.3 & 27.1   \\
        \cline{1-3}\cline{5-8}
        \multirow{6}*{F-ViT (DeCLIP)} & \multirow{3}*{ViT-B/16} &T/2-I/2 &  &25.8  &21.0  &15.1  &21.9  \\
         &  &T-I/2 &  &29.0  &21.8  &16.2  & 23.7  \\
         &  & T/2-I &  &26.4  &23.3  &23.1  & 24.5  \\
        \cline{2-3}\cline{5-8}
	   & \multirow{3}*{ViT-L/14} &T/2-I/2 &  &32.0  &33.2  &18.2  & 30.2  \\
         &  &T-I/2 &  &33.7  &34.8  &21.6  & 32.1  \\
         &  & T/2-I &  & 32.3 &33.7  &33.8  & 33.2 \\
        \cline{1-3}\cline{5-8}
        \multirow{6}*{\textbf{LVDor} (DeCLIP)} & \multirow{3}*{ViT-B/16} &T/2-I/2 &  &30.2  & 27.1 &27.3  & 28.3  \\
         &  &T-I/2  &  &31.3  &26.5  &29.7  & 28.9   \\
         &  & T/2-I &  &30.7  &27.2  &28.9  & 28.9 \\
        \cline{2-3}\cline{5-8}
         & \multirow{3}*{ViT-L/14} &T/2-I/2 &  & 37.6 & 37.1 &37.9  &37.4    \\
         &  &T-I/2   &  &38.0  &37.1  &38.9  & 37.8  \\
         &  & T/2-I  &  &37.7  &36.3  &39.1  & 37.3   \\ 
        \bottomrule    
	\end{tabular} \vspace{-15pt}
\end{table}

As shown in Table~\ref{tab:mix-LVIS}, MM-OVOD exhibits a substantial drop in overall performance, with the models trained on both datasets showing an $AP$ reduction of approximately 6\%–8\%. For F-VLM (CLIP) and F-ViT (CLIPSelf and DeCLIP), we use pre-extracted image and text prompt embeddings and obtain the final prompt through simple average-weighted fusion. Under the most challenging prompt scenario (`T/2-I/2'), these methods likewise display a clear degradation in performance.
We attribute this decline to the lack of joint modeling of complete dual-modal prompts during training, along with the overly naive averaging strategy for fusing the two modalities, which introduces significant semantic discrepancies between them. As a result, these models struggle to adapt when any modality is missing and perform poorly when confronted with flexible and heterogeneous prompt forms.

In contrast, LVDor demonstrates robustness. Under the flexible prompt setting (`T/2-I/2'), with DeCLIP ViT-B/16 and ViT-L/14, LVDor shows a 0.5\% and 0.7\% drop in $AP$, respectively, compared with the fusion results in Table~\ref{tab:LVIS}, significantly lower than the degradation of other methods. Consistent trends also appear under the other two configurations (`T-I/2' and `T/2-I'), indicating that LVDor remains stable across prompt modality combinations.
These results show that under complex and uncertain prompt configurations, LVDor effectively mitigates the effects of modality absence or inconsistency, exhibiting superior robustness and adaptability to multi-modal prompts. Consequently, it offers greater applicability and generalization for LV-OSD tasks.

As shown in the row `LVDor (CLIPSelf) w/o PRM', our experiments based on CLIPSelf ViT-B/16 systematically verify the crucial role of the Prompt Random Masking (PRM) mechanism in flexible prompt scenarios (`T/2-I/2'). The results show that removing this mechanism leads to a noticeable performance drop in multi-modal tasks, indicating that the model struggles to adapt to randomly varying prompt combinations.
This degradation primarily arises from the lack of simulated multi-modal prompt absence and random combinations during training, which prevents the model from effectively learning cross-modal interactions and limits its ability to handle incomplete or imbalanced prompt inputs during inference. Consequently, incorporating the Random Masking Mechanism substantially enhances the model's capacity to adapt to diverse prompt combinations, enabling more stable and generalizable performance under flexible prompt conditions.
Overall, the Random Masking Mechanism proves not only effective but also essential for improving the model's robustness under complex and non-fixed prompt configurations.

\setlength{\tabcolsep}{2.0pt}
    \begin{table}[t!]
    \centering
    \scriptsize
    % \footnotesize
    \caption{Ablation study of our modules on OV-LVIS.}   \vspace{-5pt}
    \label{tab:ablation}
    \begin{tabular}{c|c|c|c|c|c|c|c|c|c|c}
    \toprule
        \multirow{2}*{ID} &\multicolumn{3}{c|}{Train} & \multicolumn{3}{c|}{Test} & \multirow{2}*{$AP_f$} & \multirow{2}*{$AP_c$} & \multirow{2}*{$AP_r$} & \multirow{2}*{$AP$} \\
        \cline{2-7}
         & MPr-T & MPr-I & TPDW  & MPr-T & MPr-I & TPDW & & & & \\\hline
        \footnotesize\ding{172} &     &  &  &  &  &  & 29.1 & 21.8 & 25.3 & 25.2 \\\hline
        \footnotesize\ding{173} & \ding{51}(mean) &  &    &\ding{51}(mean)  &   &   &  29.6 & 22.7 & 25.6 & 25.9  \\\hline
        \footnotesize\ding{174} & \ding{51} &  & \ding{51}   & \ding{51} &  & \ding{51}  & 30.9 & 24.5 & 26.6 & 27.4 \\\hline
        \footnotesize\ding{175} & \ding{51} &\ding{51}  & \ding{51}   & \ding{51} &  & \ding{51} & 31.4 & 24.6 & 29.6 & 28.1 \\\hline
        \footnotesize\ding{176} & \ding{51} &\ding{51}  & \ding{51}   &      & \ding{51} & \ding{51} & 30.7 & 27.2 & 26.4 & 28.4 \\\hline
        \footnotesize\ding{177} & \ding{51} &\ding{51}  & \ding{51}   & \ding{51} & \ding{51} & \ding{51} & 31.5 & 26.8 & 29.8 & 29.2 \\
        \bottomrule 
    \end{tabular}  \vspace{-10pt}
\end{table}

\subsection{Ablation Study}
We use F-ViT (CLIPSelf ViT-B/16) as our baseline and perform the following ablation studies to examine the sensitivity of our components and hyperparameters.

\textbf{Component ablation.} In Table~\ref{tab:ablation}, we conduct ablation studies on our proposed multi-modal prompts (MPr), and the Target-guided Prompt Dynamic Weighting (TPDW) module.

Row \ding{172} indicates the use of a standard template for training a detector based on text modality prompts (F-ViT).

Row \ding{173} refers to training the detector with the mean of the textual descriptions from our proposed MPr (MPr-T), achieving accuracies of 29.6\%, 22.7\%, 25.6\%, and 25.9\% for the `frequent', `common', `rare', and overall categories, respectively. These results represent improvements of 0.5\%, 0.9\%, 0.3\%, and 0.7\% compared to the baseline. This suggests that detailed appearance attribute descriptions (generated by MPr) can provide better transfer support for CLIP in downstream tasks, effectively preserving its generalization ability through enriched semantic information.

Row \ding{174} demonstrates training with textual descriptions from MPr and further integrating TPDW, followed by inference using the same approach. 
This method further enhances the results compared to row \ding{173}, with accuracy improvements of 1.3\%, 1.8\%, 1.0\%, and 1.5\% in the `frequent', `common', `rare', and overall categories, respectively. This is because the TPDW strategy, which dynamically adjusts prompts based on the input image, contributes to better preservation of CLIP's semantic alignment and generalization capabilities.

Rows \ding{175} \ding{176}, and \ding{177} further integrate the image modality (MPr-I) during training, which is the whole framework for training of our method. 
Compared to raw  \ding{174}, we can see the performance improvement since the use of both modalities during training allows for more detailed and flexible semantic representation.
Comparing Rows \ding{175} \ding{176} and \ding{177}, we can see that, during testing, using the single modality of text, image, performs slightly poorly than both of them. But, all the results are acceptable, which, yet again, demonstrates that our method is robust to the input in the real applications.

\textbf{Sensitivity to the number of feature patches.} As shown in Table~\ref{tab:patch}, we conduct a sensitivity analysis on the number of patches used to divide the full-image feature map in Sec.~\ref{sec:TPDW}. We set the patch numbers to 1, 4, and 9, yielding AP scores of \%, \%, and \%, respectively. Empirically, using 4 patches proves sufficient, as increasing the number of image queries provides no additional benefit. Therefore, we adopt four image queries as the default setting throughout all experiments.

\begin{table}[t!]
    \centering
    \scriptsize
    % \tiny
    \caption{Sentitivity Analysis of the number of feature patches.}\vspace{-5pt}
    \label{tab:patch}
    \begin{tabular}{c|c|ccc}
    \toprule
    % \textbf{$#Patches$} & Modality & 1  & 4  & 9 \\
    \textbf{\# Patches} & Modality & 1  & 4  & 9 \\
    \midrule
    \textbf{$\textit{AP}$} &F   &28.0 (16.75 FPS)    &29.2 (13.61 FPS)    & 28.8 (8.33 FPS)    \\
    \bottomrule
    \end{tabular}\vspace{-15pt}
\end{table}

\textbf{Sensitivity to top $K$ candidate categories.} As shown in Table~\ref{tab:k}, we conduct a sensitivity analysis on the number of candidate categories (top $K$) used in Sec.~\ref{sec:TPDW}. We vary $K$ across 1, 3, 5, 7, and 9, and observe that using five candidate categories strikes a favorable balance—achieving competitive AP performance while keeping computational overhead moderate. Therefore, we adopt $K=5$ as the default setting in all experiments.

\begin{table}[t!]
    \centering
    \tiny
    \caption{Sensitivity Analysis of the top $K$ candidate categories.}
    \vspace{-5pt}
    \label{tab:k}
    \resizebox{\linewidth}{!}{
    \begin{tabular}{c|c|ccccc}
    \toprule
    \textbf{$K$} & Modality  & 1  & 3  & 5  & 7  & 9 \\
    \midrule
    \textbf{$\textit{AP}$} & F  
    & 28.0 (16.67 FPS)  
    & 28.7 (15.12 FPS)  
    & 29.2 (13.61 FPS)  
    & 29.2 (12.10 FPS)  
    & 28.9 (10.32 FPS)   \\
    \bottomrule
    \end{tabular}
    }
    \vspace{-15pt}
\end{table}

\subsection{Cross-Dataset Transfer Results}
Table~\ref{tab:O365} presents the cross-dataset transfer results of our method from OV-LVIS to Objects365~\cite{shao2019objects365}. We compare our method with Detic, MM-OVOD, F-VLM (CLIP), and F-ViT (CLIPSelf and DeCLIP), reporting the bounding box $AP$ metric following the Objects365 standard. Among these methods, Detic and MM-OVOD are trained on LVIS-all, and their variants further employ IN-L as additional weak supervision, whereas the other methods use models trained only on the LVIS base categories (LVIS-base) for evaluation. All trained detectors are evaluated on the Objects365 validation set. Following the setting in MM-OVOD, we treat the least frequent $\frac{1}{3}$ of categories in Objects365 as rare categories.

After training on LVIS-base, we follow the method outlined in Section `MPr' to generate textual descriptions for each category in the Objects365. We then gather example images from the internet for each category and encode them using a frozen CLIP encoder to serve as multi-modal prompts. 
It can be observed that under the text‐prompt setting, our method achieves consistent and notable performance gains over F-VLM (CLIP) and F-ViT (CLIPSelf and DeCLIP). We further evaluate the model with image prompts and fused prompts, where the fusion setting delivers the best results, fully aligning with the conclusions in Table~\ref{tab:LVIS}.
These findings demonstrate that our approach effectively leverages high-quality category descriptions generated by large language models, using them as text prompts to provide the detector with more fine-grained and discriminative semantic priors than those derived from fixed template-based prompts. Moreover, when incorporating the image modality and fusing it with text, the model aligns more easily with cross-dataset visual semantic patterns and structures, resulting in stronger robustness and generalization during cross-dataset transfer. Overall, combining fused prompts with our proposed framework enables the model to maintain superior detection performance in open-set scenarios while exhibiting remarkable cross-dataset transferability.

\begin{table}[t!]
	\centering
	\scriptsize
    % \footnotesize
    % \tiny
    \caption{Cross-dataset main results on Object365.} \vspace{-5pt}
	\label{tab:O365}
	\begin{tabular}{l|c|c|c|c|c|c}
        \toprule
        Method  &Backbone   & Modality   & Train Data   &$AP_r$    &$AP$   &$AP50$\\\hline
        \multirow{2}*{Detic}   & \multirow{4}*{{RN50}$^*$}     & \multirow{4}*{T}     & LVIS-all    & 9.5   & 13.9  &19.7  \\
           &    &  & LVIS-all + IN-L   &12.4    & 15.6    &22.2  \\
        \cline{1-1}\cline{4-7}
        \multirow{2}*{MM-OVOD} &  &    & LVIS-all   & 10.1    & 14.8     &21.0    \\
         &  &  & LVIS-all + IN-L   & 13.1 & 16.6  & 23.1   \\
        \hline

    	\multirow{2}*{F-VLM (CLIP)} & RN50 & \multirow{2}*{T}   & \multirow{24}*{LVIS-base}  &-  &11.9  &19.2    \\
    	   & RN50×4 &  &  &-   &14.2  &22.6   \\
        \cline{1-3}\cline{5-7}
        \multirow{6}*{\textbf{LVDor} (CLIP)} & \multirow{3}*{RN50} & T  &  &11.4   &15.2  &23.7   \\
         & & I   &  &11.0  &13.7  &22.4      \\
         & & F   &  &11.9  &15.6  &23.9      \\
         \cline{2-3} \cline{5-7}
         & \multirow{3}*{RN50×4} & T  &  &14.8  &17.9  &26.7  \\
         & & I   &  &14.2  &17.5  &26.3        \\
         & & F   &  &15.0  &18.1  &26.9    \\
        \cline{1-3}\cline{5-7}
    	\multirow{2}*{F-ViT (CLIPSelf)} & ViT-B/16 & \multirow{2}*{T}   &   &16.8  & 19.0  &32.3    \\
    	   & ViT-L/14 &  &   &21.7  &23.7  &39.2  \\
        \cline{1-3}\cline{5-7}
        \multirow{6}*{\textbf{LVDor} (CLIPSelf)} & \multirow{3}*{ViT-B/16} & T  &  &18.5  &20.1  &33.9    \\
         & & I   &  &16.5  &19.2  &32.8      \\
         & & F   &  &19.1  &20.4  &34.3     \\
         \cline{2-3} \cline{5-7}
         & \multirow{3}*{ViT-L/14} & T  &  &23.1  &24.9  &40.9      \\
         & & I   &  &22.5  & 23.6  &39.4       \\
         & & F   &  &23.8  &25.2  &41.4    \\
        \cline{1-3}\cline{5-7}
        \multirow{2}*{F-ViT (DeCLIP)} & ViT-B/16 & \multirow{2}*{T}   &   &17.6  &20.2  &33.1    \\
    	   & ViT-L/14 &  &   &22.3  &24.5  &39.8    \\
        \cline{1-3}\cline{5-7}
        \multirow{6}*{\textbf{LVDor} (DeCLIP)} & \multirow{3}*{ViT-B/16} & T  &  &19.2  &20.6  &34.5   \\
         & & I   &  &17.1  &19.6  &33.0        \\
         & & F   &  &19.6  &20.9  &34.9      \\
         \cline{2-3} \cline{5-7}
         & \multirow{3}*{ViT-L/14} & T  &  &23.7  &25.4  &41.2       \\
         & & I   &  &22.7  &23.8  &39.8        \\
         & & F   &  &24.3  &26.1  &42.0     \\
	\bottomrule    
	\end{tabular} \vspace{-15pt}
\end{table}

\section{Inference Efficiency Analysis}
We evaluate the inference speed of all methods on a single NVIDIA 3090 GPU. F-ViT (ViT-B/16) reaches 21.18 FPS, while LVDor with dual-modal prompts achieves 13.61 FPS. Although the TPDW module introduces some computational overhead, the resulting improvement in detection performance makes this cost a reasonable and acceptable trade-off in open-set detection.
To address the common lack of inference efficiency reporting in existing methods, we further benchmark MM-OVOD~\cite{kaul2023multi} and OV-DQUO~\cite{wang2025ov}, which attain 2.09 FPS and 2.31 FPS, respectively. In comparison, LVDor strikes a far more competitive balance between accuracy and efficiency, significantly outperforming other approaches and demonstrating stronger potential for practical deployment.

\section{Visualization results}
\begin{figure*}[t!]
	\centering
	\includegraphics[width=0.95\linewidth]{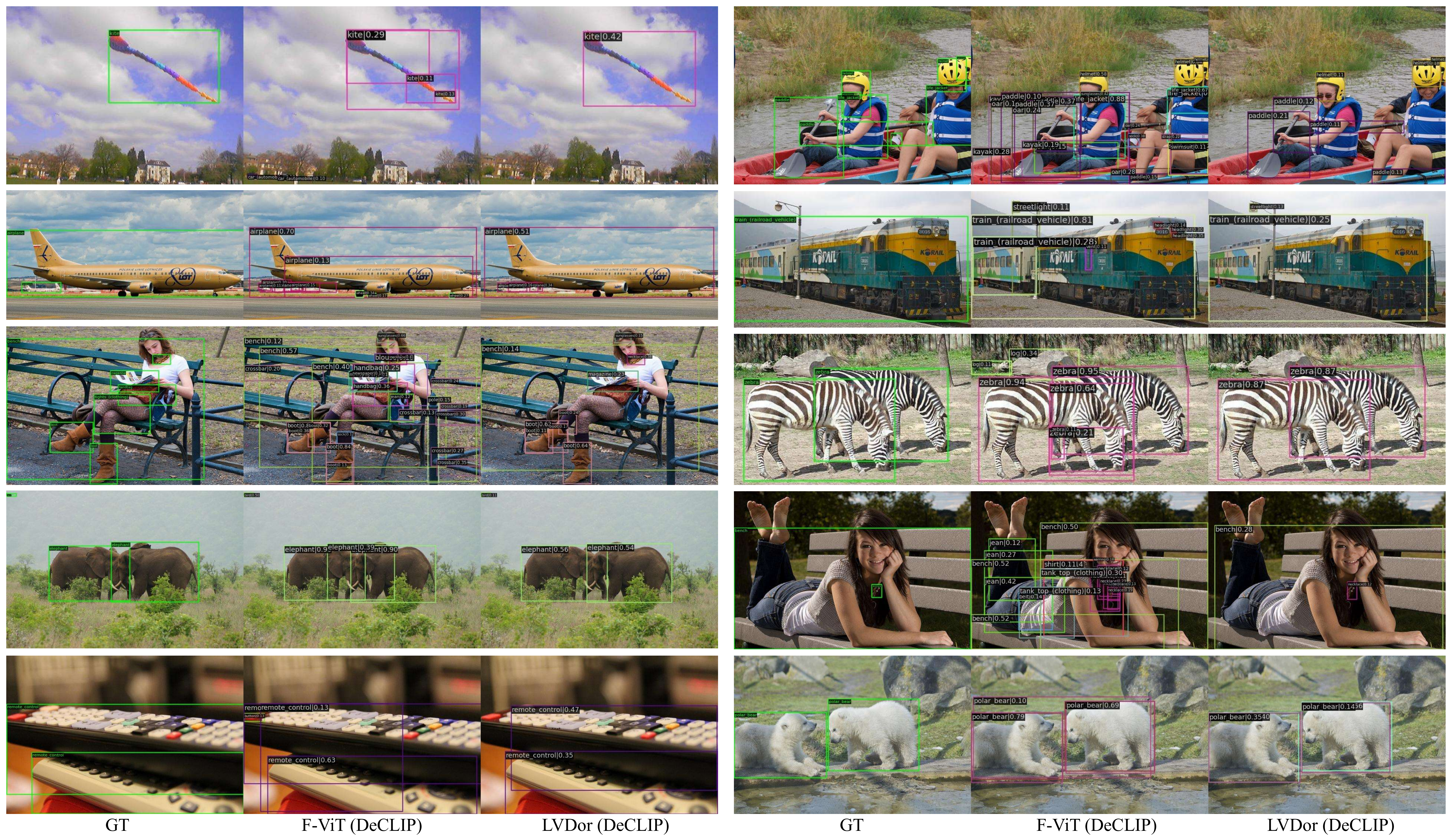}
	\vspace{-10pt}
	\caption{Visualization of detection results.}
	\label{fig:results2}
	\vspace{-18pt}
\end{figure*}

As shown in Fig.~\ref{fig:results2}, we present qualitative results for the open-set object detection task on the OV-LVIS dataset based on complementary language-vision prompting. The figure displays three types of outputs: ground-truth annotations on the left, predictions from F-ViT (DeCLIP ViT-B/16) using text-only prompts in the center, and results from our proposed LVDor (DeCLIP ViT-B/16) with fused dual-modal prompts on the right.
From the visual comparison, it is evident that LVDor exhibits markedly superior performance in both localization precision and category recognition accuracy compared with F-ViT. Our method identifies semantically relevant instances more accurately while effectively suppressing background noise and false positives, producing more reliable and semantically coherent detections. Notably, LVDor maintains strong performance even in fine-grained categories or complex scenes, highlighting the synergistic benefits of dual-modal prompting in enhancing visual–semantic understanding.
These qualitative results further validate the robustness and strong generalization ability of our method in challenging open-set scenarios, demonstrating that fusing language and visual prompts provides richer semantic guidance and significantly improves overall open-set object detection performance.

\section{Examples of Textual Descriptions and Image Queries}  
\label{sec:examples}
\begin{figure}[h!]
	\centering
	\includegraphics[width=0.98\linewidth]{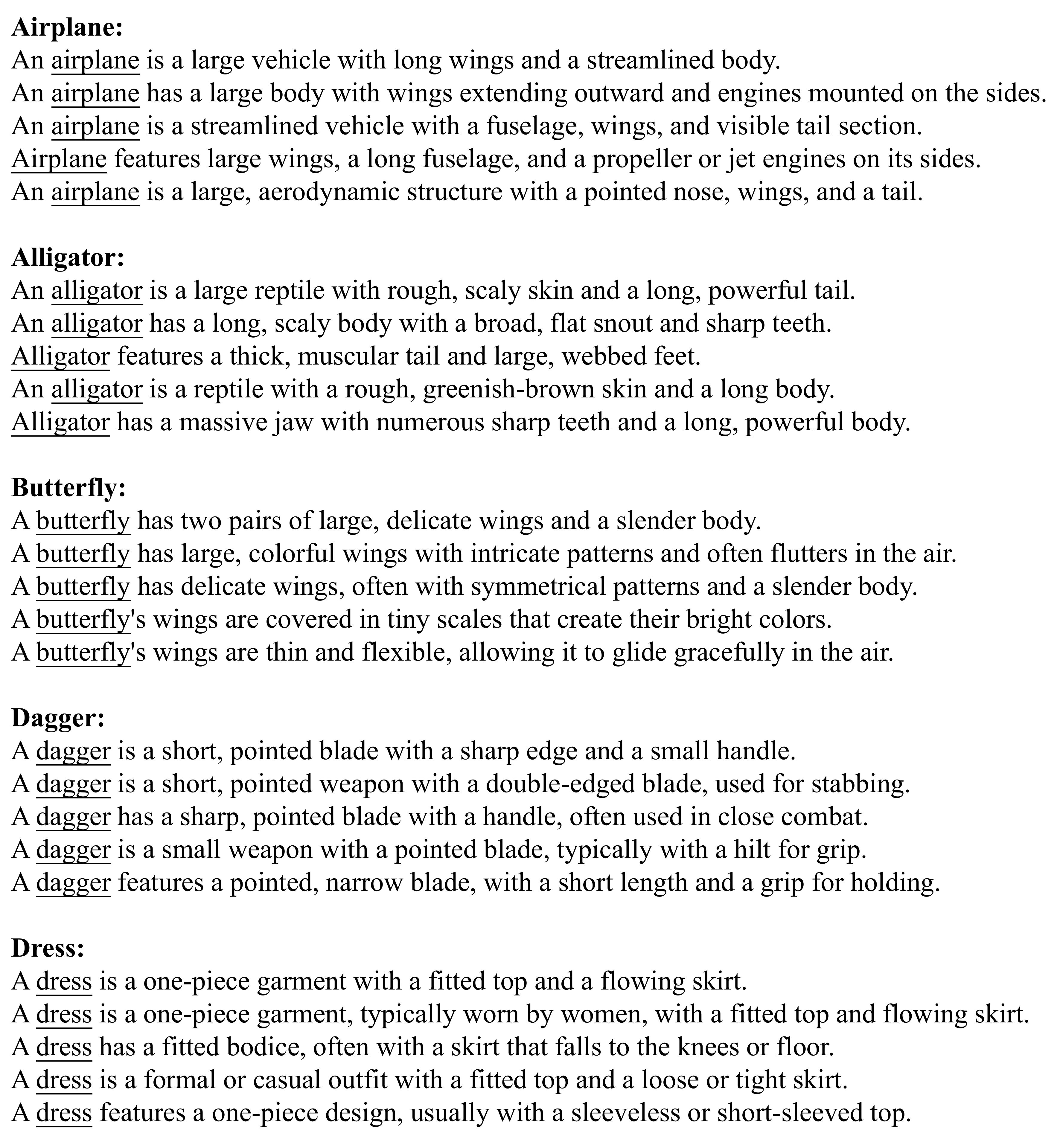}
	\vspace{-10pt}
	\caption{Category descriptions of the OV-LVIS dataset.}
	\label{fig:text prompts}
	\vspace{-10pt}
\end{figure}

\begin{figure}[h!]
	\centering
	\includegraphics[width=0.95\linewidth]{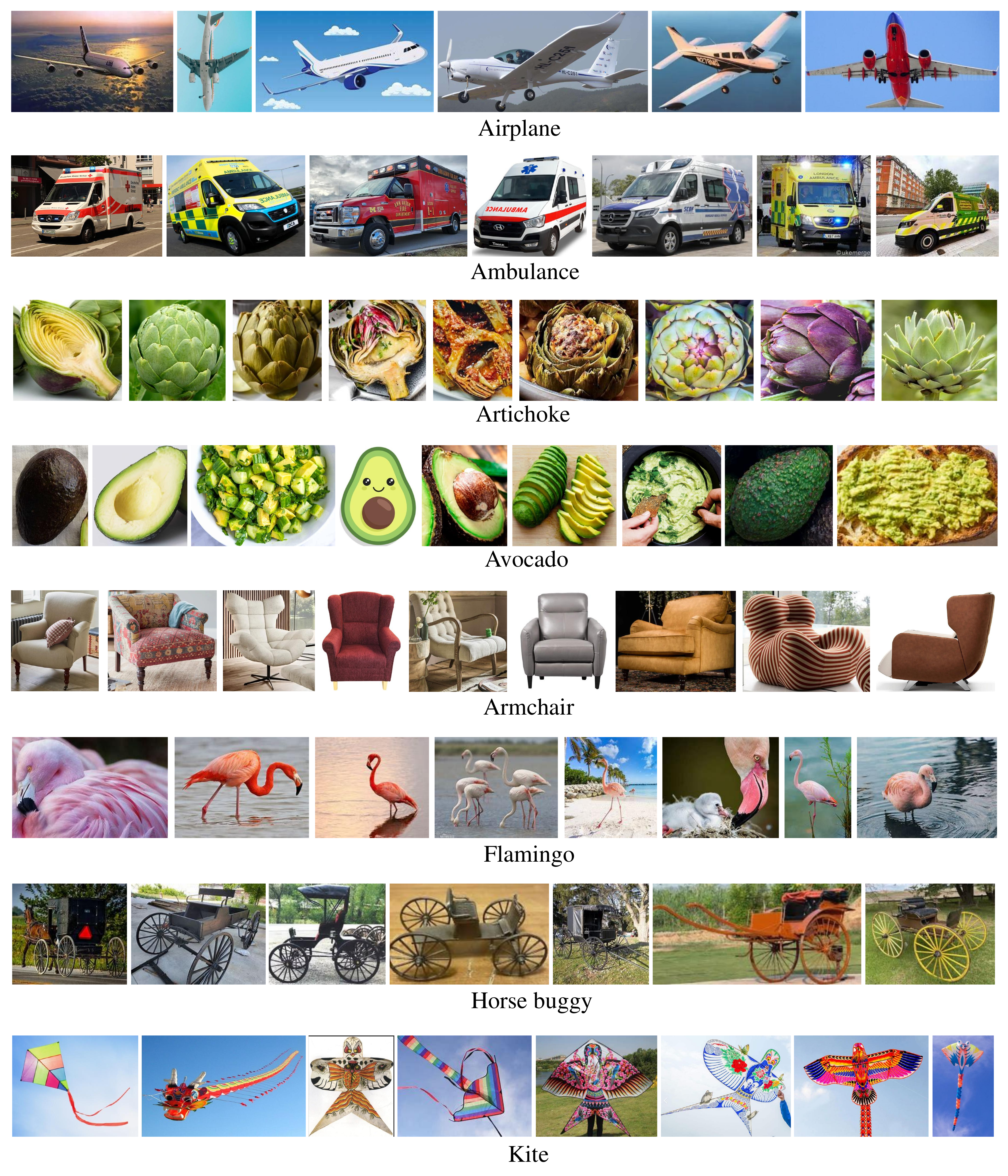}
	\vspace{-10pt}
	\caption{Image examples for each category of the OV-LVIS dataset.}
	\label{fig:image prompts}
	\vspace{-15pt}
\end{figure}

As shown in Fig.~\ref{fig:text prompts}, we use Deepseek to generate five semantically rich and fine-grained textual descriptions for each category in the OV-LVIS dataset, covering appearance traits, semantic attributes, and potential variations. The complete generation process is detailed in the subsection \textit{Text-based category description generation}. Meanwhile, Fig.~\ref{fig:image prompts} presents diverse image examples collected for each category in the subsection \textit{Image-based category example collection}, encompassing different viewpoints, scales, shapes, and scene conditions, thereby providing more representative visual priors for multi-modal prompting.

\section{Conclusion}
\label{sec:conclusion}

In this work, we have proposed a novel approach to open-set object detection based on language-vision complementary prompts. 
By constructing the Multi-modal Prompts (MPr) that integrate both text and image prompts, our method overcomes the limitations of traditional single-modality approaches, significantly enhancing the model's flexibility. 
The introduction of the Target-guided Prompts Dynamic Weighting (TPDW) method further strengthens the model's ability to dynamically select the optimal prompts based on specific scenarios. 
Through extensive experiments on OV-LVIS and Object365, we validate the effectiveness of our approach, achieving SOTA performance. 
Our work provides a more flexible and practical solution for object detection tasks in real-world applications.

\section{Limitations and Future Directions}
Although LVDor shows strong performance and flexibility under the LV-OSD setting, it still has inherent limitations. These mainly arise from its reliance on CLIP, whose representational capacity is bounded by the underlying backbone. To balance generalization and training efficiency, we use a frozen vision–language backbone; however, this may limit further gains under severe domain shifts and still falls short of real-time inference. Future work will explore lightweight, zero-shot–oriented designs for LVDor and extend the proposed language–vision complementary paradigm to broader open-set and dense perception tasks.

{
\small
\bibliographystyle{unsrt}
\bibliography{ref}
}

\end{document}